\documentclass[letterpaper, 10 pt, conference]{ieeeconf}  

\IEEEoverridecommandlockouts                              

\usepackage{amsmath}
\usepackage{amssymb}

\usepackage{bm}
\usepackage{amsfonts}
\let\labelindent\relax
\usepackage[inline]{enumitem}
\usepackage{graphicx}
\usepackage{multirow}
\usepackage{booktabs} 
\usepackage{algorithm}[H]
\usepackage[noend]{algpseudocode}%

\usepackage{booktabs}
\usepackage{threeparttable}

\usepackage{hyperref}
\usepackage{microtype}

\title{\LARGE \bf
Dynamics-Aware Quality-Diversity for Efficient Learning of Skill Repertoires
}

\author{Bryan Lim$^{1}$, Luca Grillotti$^{1}$, Lorenzo Bernasconi$^{1}$, and Antoine Cully$^{1}$
\thanks{$^{1}$Imperial College London, United Kingdom.
        {\tt\small \{bwl116, lg4615, ljb818, a.cully\}@ic.ac.uk}}%
}

\begin{document}

\maketitle
\thispagestyle{empty}
\pagestyle{empty}

\newcommand{\algonamefull}{Dynamics-Aware Quality-Diversity}
\newcommand{\algoname}{DA-QD}
\renewcommand{\vec}[1]{{\boldsymbol{{#1}}}} 
\newcommand{\mat}[1]{{\boldsymbol{{#1}}}} 

\newcommand{\param}[1]{\vec \phi_{{#1}}}
\newcommand{\parammutated}[1]{\vec \phi^{\ast}_{{#1}}}
\newcommand{\paramsynthetic}[1]{\smash{\widetilde{\vec \phi_{{#1}}}}}
\newcommand{\paramsyntheticmutated}[1]{\widetilde{\vec \phi^{\ast}_{{#1}}}}

\newcommand{\policy}[1]{\pi_{{#1}}}
\newcommand{\policyreal}[1]{\policy{\param{{#1}}}}
\newcommand{\policyrealmutated}[1]{\policy{\parammutated{{#1}}}}

\newcommand{\policysynthetic}[1]{\policy{\paramsynthetic{{#1}}}}
\newcommand{\policysyntheticmutated}[1]{\policy{\paramsyntheticmutated{{#1}}}}

\newcommand{\batchpoliciesreal}[1][b]{\policyreal{1} \ldots \policyreal{{#1}} }
\newcommand{\batchpoliciesrealmutated}[1][b]{\policyrealmutated{1} \ldots \policyrealmutated{{#1}} }

\newcommand{\batchpoliciessynthetic}[1][b]{\policysynthetic{1} \ldots \policysynthetic{{#1}} }
\newcommand{\batchpoliciessyntheticmutated}[1][b]{\policysyntheticmutated{1} \ldots \policysyntheticmutated{{#1}} }

\newcommand{\archivereal}{\mathcal{A}}
\newcommand{\archivesynthetic}{\smash{\widetilde{\mathcal{A}}}}

\newcommand{\skilldescriptor}{\vec{sd}}
\newcommand{\reward}{R}

\newcommand{\dynamicsmodel}{\widetilde{p}_{\vec \theta}}
\newcommand{\replaybuffer}{\mathcal{B}}

\newcommand{\outcomespace}{\mathcal{O}}

\newcommand{\outcomegp}{\vec o}

\begin{abstract}

Quality-Diversity (QD) algorithms are powerful exploration algorithms that allow robots to discover large repertoires of diverse and high-performing skills. However, QD algorithms are sample inefficient and require millions of evaluations. In this paper, we propose Dynamics-Aware Quality-Diversity (DA-QD), a framework to improve the sample efficiency of QD algorithms through the use of dynamics models. We also show how \algoname{} can then be used for continual acquisition of new skill repertoires.
To do so, we incrementally train a deep dynamics model from experience obtained when performing skill discovery using QD. We can then perform QD exploration in imagination with an imagined skill repertoire. 
We evaluate our approach on three robotic experiments. First, our experiments show \algoname{} is 20 times more sample efficient than existing QD approaches for skill discovery.
Second, we demonstrate learning an entirely new skill repertoire in imagination to perform zero-shot learning. Finally, we show how \algoname{} is useful and effective for solving a long horizon navigation task and for damage adaptation in the real world. 
Videos and source code are available at: \url{https://sites.google.com/view/da-qd}.

\end{abstract}


\section{INTRODUCTION} \label{sec:introduction}

Despite recent progress, deploying a robot in open-ended real-world settings remains a challenge. This is mostly due to the difficulty for engineers to anticipate all the situations that the robot might have to face. In this context, employing learning algorithms with strong exploration capabilities represents a promising way to enable robots to discover new skills and find new ways to deal with unexpected situations. On the other hand, most exploration methods are sample inefficient and rely on the use of simulators. However, once robots are deployed, sample efficiency is important when acting in the real world.
To be effective and to continually learn when deployed in open-ended real-world environments, robots need sample-efficient exploration algorithms.

In reinforcement learning, Quality-Diversity (QD) algorithms~\cite{cully2017quality,pugh2016quality} have been shown to be powerful exploration algorithms that can solve environments with deceptive rewards and previously unsolved hard exploration tasks on the Atari and robotic domains~\cite{ecoffet2021first}. The importance of QD for exploration in open-ended real-world environments is twofold. Firstly, QD produces collections of high-performing and diverse skills which can be leveraged for versatility and adaptation in real-world environments~\cite{cully2015robots, chatzilygeroudis2018reset, kaushik2020fast}. Secondly, QD is also important during the exploration process. Simultaneously optimizing and preserving a repertoire of policies to diverse skills (or goals) allows for \textit{goal switching}~\cite{nguyen2016understanding, clune2019ai} which enables policies from different skills (or goals) to act as stepping stones to higher performing policies. However, QD algorithms are inherently sample inefficient and heavily leverage the use of simulators. 

A common solution to improve sample-efficiency is to use model-based learning methods~\cite{wang2019benchmarking, chua2018deep, kurutach2018model, clavera2018model}. They can incorporate off-policy data and generic priors, making them highly sample efficient. The learnt models can also be re-purposed to solve new tasks~\cite{sekar2020planning}. However, the diversity of training data affects the accuracy of the learned model~\cite{sekar2020planning}. Prior work either learns forward models using random policies~\cite{ha2018world} or update and refine the dynamics model incrementally as the policy is being optimized for the task~\cite{chua2018deep, nagabandi2020deep, janner2019trust, hafner2019learning, hafner2019dream}.
Instead of random or task-specific policies, our work learns a dynamics model using the diverse policies obtained with QD during skill discovery.

In this paper, we introduce Dynamics-Aware Quality Diversity (\algoname{}), an algorithm that combines deep dynamics models with QD to efficiently perform skill discovery. We use the models to seek out expected novel policies in imagination during skill discovery with QD. Using dynamics models in QD benefits both the QD exploration process and model learning. This lets us learn novel and diverse skills using QD purely from imagined states without any or minimal environment interactions, increasing the sample efficiency of QD by an order of magnitude. At the same time, the QD exploration process inherently provides a rich and diverse dataset of transitions which enables better models to be learnt.





The key contributions of this paper are:
\begin{itemize}[noitemsep, nolistsep, leftmargin=5mm]
\vspace{-2mm}
    \setlength\itemsep{0.1em}
    \item \textbf{Framework to improve the sample-efficiency of QD using dynamics models.} We show that using dynamics models and the introduction of an imagined skill repertoire enable \algoname{} to be about 20 times more sample-efficient than previous QD methods for skill discovery. 
    
    \item \textbf{Continual acquisition of new skills.} We demonstrate re-use of the dynamics models for QD skill discovery. QD can be performed purely in imagination to learn new skills in a few-shot or zero-shot manner. This was not possible before with prior QD methods.
        
   \item \textbf{Real-world navigation and adaptation capabilities within low evaluation budget.} For long horizon tasks, we pair \algoname{} with an existing skill planning method. Skill repertoires discovered using \algoname{} outperform baseline skill repertoires generated with equivalent evaluation budget and are equivalent to methods using a significantly larger budget. We show this by performing extensive real-world experiments on navigation tasks, in both damaged and undamaged conditions.

\end{itemize}


\begin{figure*}[t]
\centering
	\includegraphics[width=0.8\textwidth]{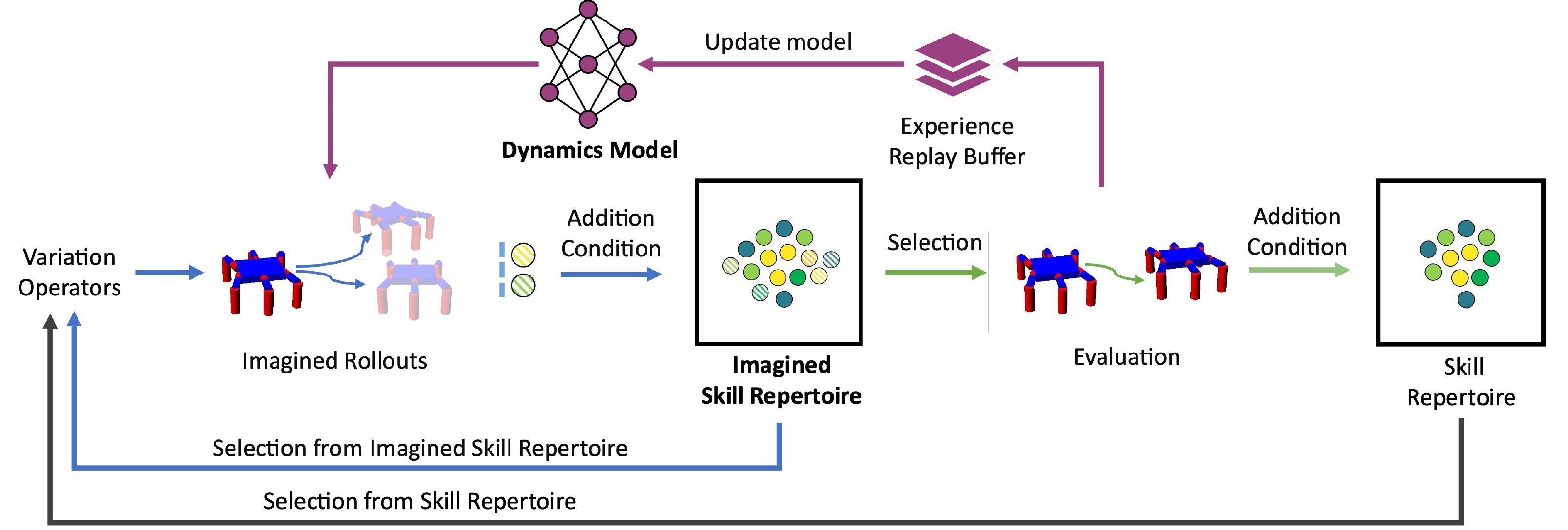}
\centering

\caption{
\textbf{Components of \algoname{}.} 
Top (Purple): The agent learns a dynamics model from a dataset of diverse experience generated from skill discovery using QD. 
Left (Blue): \algoname{} uses the dynamics model to perform imagined rollouts to predict the return and skill descriptor and build an imagined repertoire. Depending on the situation, the agent can choose to continue to perform skill discovery using QD with the imagined repertoire or act in the environment.
Right (Green): Agent acts in the environment by selecting skills discovered in its imagined repertoire.
Depending on the outcome, the skills may be added to the repertoire.
\algoname{} returns a repertoire of diverse and high-performing skills; this repertoire can then be used with planning methods.
}
\vspace{-4mm}
\label{fig:mbme-framework}
\end{figure*}

\section{BACKGROUND: Quality-Diversity} \label{sec:background}
Contrary to conventional single-objective optimization algorithms that search for a single performance-maximizing solution, Quality-Diversity (QD) algorithms generate a collection of diverse and high-performing solutions. QD algorithms achieve this using the notion of novelty which distinguishes between solutions with different behaviours. This novelty is what drives the search and exploration process. To compute novelty, most QD algorithms maintain a repertoire that contains only novel and the highest-performing solutions encountered during the search process. The repertoire is also the final output of the algorithm.
The novelty of a policy is estimated via a skill descriptor, denoted $\skilldescriptor$.
The skill descriptor is a numerical vector that characterizes the behaviour of a policy.
QD algorithms learn a repertoire of policies that explore and cover the skill descriptor space as much as possible. 
For example, if we want a robot to discover locomotion primitives that allow it to reach any location around it, the final position of the robot can be used as a skill descriptor~\cite{cully2013behavioral, chatzilygeroudis2018reset}. 

MAP-Elites~\cite{mouret2015illuminating} is a simple QD algorithm that discretizes the skill space into a grid. Different variations of MAP-Elites manage the repertoire differently~\cite{vassiliades2016using, fontaine2019mapping}. Our approach builds on MAP-Elites but uses an unstructured repertoire~\cite{lehman2011evolving, cully2017quality}, as its flexibility allows the density of the skill space to be adapted easily and enables further extensions such as unsupervised skill descriptors~\cite{cully2019autonomous}. 

The skill repertoire $\archivereal$ begins empty and is initialized with a set of randomly generated policies, sampled in the parameter space. QD then follows a loop that; (1) selects and copies $N$ policies $(\policyreal{i})_{i\in\left[1..N\right]}$ from the repertoire $\archivereal$, (2) applies a variation operator on the parameters $\param{i}$ of the selected policies to get new parameters $\parammutated{i}$, (3) evaluates each newly formed policy $\policyrealmutated{i}$ and gathers the agent trajectory to obtain the skill descriptor $\skilldescriptor_i$ and the sum of rewards (return) $\reward_i$, and (4) updates the repertoire based on the novelty score and the return of the policies. The novelty score of a policy $\policy{}$ is computed as the average Euclidean distance to its $K$ nearest skill descriptors $(\skilldescriptor_k)_{k\in \left[1..K\right]}$ from the repertoire:
\vspace{-4pt}\begin{equation}
\label{equation:novelty-score}
    novelty(\policy{}) = \frac{1}{K}\sum_{k=1}^{K} \|\skilldescriptor_{\policy{}} - \skilldescriptor_k \|_2
\end{equation}
We use the same repertoire addition condition as in Cully et al., 2017~\cite{cully2017quality}. A policy is added to the repertoire if (i) the distance to its nearest neighbour exceeds a predefined threshold, $l$, meaning it is novel enough. If the distance to its nearest neighbour is lower than $l$, the skill can replace its nearest neighbour in the repertoire if: (ii) the distance to the second nearest neighbour is greater than $l$ and if it improves the repertoire either by having a greater quality score or novelty score over its nearest neighbour. Otherwise, this policy will be discarded. These repertoire addition conditions ensure both quality (exploitation) in the form of local competition within a certain niche of skills, and diversity (exploration) through the preservation of novel behaviours in our skill repertoire.

Traditionally, QD search is driven by an Evolutionary Algorithm (EA) for its effectiveness in diversifying the search to find novel skills. In our work, we use uniform selection operators and EA variation operators, which perturb the parameters of a policy during the optimization process to generate new policies. Specifically, we use the directional variation~\cite{vassiliades2018discovering}:
\vspace{-2pt}
\begin{equation}
\label{equation:directional-variation}
    \parammutated{} = 
    \param{1} 
    + \sigma_1 \mathcal{N}(0, \mat{I}) 
    + \sigma_2 \mathcal{N}(0, 1) (\param{2} - \param{1})
\end{equation}
where $\param{1}$ and $\param{2}$, are the parameters of two randomly selected skill policies from the repertoire. Updated parameters, $\parammutated{}$, are obtained by adding Gaussian noise with a scalar covariance matrix $\sigma_1 N(0, \mat I)$ to $\param{1}$ and displacing the parameter vector along the line from $\param{1}$ to $\param{2}$. 


\section{RELATED WORK}\label{sec:related-work}
\textbf{Skill Discovery for Adaptation.} 
Quality-Diversity (QD) algorithms~\cite{pugh2016quality, cully2017quality, lehman2011evolving} can be used to learn a large repertoire of diverse and high-performing skills~\cite{cully2013behavioral}. Such repertoires have allowed for rapid damage recovery for robots~\cite{cully2015robots, chatzilygeroudis2018reset, colas2020scaling, kaushik2020adaptive} and can also be used by a planning algorithm to perform more meaningful longer-horizon tasks~\cite{chatzilygeroudis2018reset}. Our work builds on QD-based methods for skill discovery to improve the sample-efficiency bottleneck of these methods.

Other methods for skill discovery exist. These are commonly based on maximizing mutual information (MI) between states in a trajectory and a sampled latent code~\cite{gregor2016variational, eysenbach2019diayn, hansen2019fast, sharma2019dynamics}. The latent variable used in these methods is analogous to the skill descriptor used in QD-based methods. Despite some similarities and having the same goal of obtaining a diverse set of policies (skills), these methods differ in their fundamentals. MI maximization methods represent the set of policies as a single latent-conditioned policy while QD methods explicitly keep a large number of individual policies to form an archive. 
As a result of this choice of representation, MI-maximization methods optimize the policy using gradient-based RL algorithms driven by the MI objective as an intrinsic reward.
On the other hand, QD methods rely on incrementally building a repertoire to maximize diversity in the skill descriptor space and locally optimize policies.
This mechanism enables QD methods to consider task-based rewards during skill-discovery, which is not trivial in MI-maximization methods~\cite{kumar2020one}. 
Due to the core differences between these methods, it would not be fair nor straightforward to compare them with the same evaluation metrics and ablations. Additionally, given that our focus is on the improvement of QD algorithms, we do not compare with MI maximization methods for skill-discovery.


\textbf{Models and Sample Efficiency in QD Search.} 
Evolution-based methods typically require a large number of evaluations during the optimization process. The sample efficiency of such evolutionary algorithms can be improved via the use of surrogate models. Such models approximate the reward function and can be used in place of performing expensive evaluations. Inspired by this, SAIL~\cite{gaier2018data} introduced the use of a surrogate model for QD search and showed an increase in sample efficiency when evaluating aerodynamic designs. However, SAIL cannot be used to predict the skill descriptor. 

Closest to our work, M-QD~\cite{keller2020model} uses a neural network as a surrogate model to predict both reward and skill descriptor. M-QD directly models the mapping from the parameter space to the skill descriptor space and the return function. We will refer to this as a \textit{direct model}. This model can also be used to adapt to skills in-between the discrete skills found in a repertoire. Like M-QD, we learn a model that can be used as a surrogate model to predict both the return and skill descriptor, but unlike M-QD our model is not task-specific and can be re-used to learn additional skill repertoires. 
An alternative method that has been shown to increase the sample efficiency of QD is the addition of more complex optimization techniques.
When combined with policy gradients~\cite{nilsson:hal-03135723} and evolution strategies~\cite{colas2020scaling, fontaine2020covariance, cully2020multi}, the efficiency of QD algorithms can be improved.
Our work is complementary to this line of work and these methods can easily be incorporated into our framework in future work.



\textbf{Model-based Reinforcement Learning.}
Model-based RL (MBRL) methods are effective methods for efficient learning. Recent methods involve learning forward models using deep neural networks which are capable of capturing complex systems dynamics~\cite{nagabandi2020deep}. Methods then fall into two main categories; i) model-based policy search methods that alternate between updating the dynamics model and learning a policy using imagined rollouts from the model's predictions~\cite{hafner2019dream, janner2019trust, rajeswaran2020game}, and ii) model-based planning methods which generally use online planning algorithms with the model to choose optimal actions in a model predictive control (MPC) loop~\cite{chua2018deep, nagabandi2020deep}. Our work is closer to the first setting. We alternate between incrementally learning a deep dynamics model and performing QD search. However, instead of optimizing a single policy using the model, we learn and optimize a repertoire of diverse policies.

Another common choice of transition model in MBRL is Bayesian models such as Gaussian Processes (GPs). GPs have shown to be effective and remarkably sample efficient especially when there are smaller amounts of data and for low-dimensional problems~\cite{deisenroth2011pilco}. However, GP models do not perform well with high-dimensional data and suffer from ill-suited kernels on tasks in which dynamics are complex~\cite{calandra2016manifold}. In our work, we leverage the strengths of GPs by using them on a higher level of abstraction: the GPs operate on the skill descriptor space, which is both low-dimensional and has fewer discontinuities.

\section{Dynamics-Aware Quality-Diversity}
\algoname{}, illustrated in Figure~\ref{fig:mbme-framework}, is an extension to QD algorithms that incorporates a dynamics model to increase sample efficiency of QD and allows for the continual acquisition of new skills. 
\algoname{} introduces two new components over the standard QD framework: a dynamics model $\dynamicsmodel$, and an imagined skill repertoire $\archivesynthetic$.
We learn a dynamics model using experience from skill discovery with QD (\ref{sec:methods-dyn_model}).
Using the dynamics model, we perform QD exploration in imagination and build an imagined skill repertoire (\ref{sec:methods-qd_exploration_imag}). Finally, we select policies from the imagined repertoire; these policies are executed to build the final skill repertoire (\ref{sec:methods-acting_in_env}).
Only evaluating policies novel or high-performing enough to be added to the imagined repertoire enables \algoname{} to be significantly more data-efficient.

\subsection{Learning Dynamics Models} \label{sec:methods-dyn_model}
\algoname{} learns a forward dynamics model $\dynamicsmodel(\vec s_{t+1} | \vec s_t, \vec a_t)$, and recursively uses the model predictions to imagine trajectories. 
We use an ensemble of probabilistic neural networks as the forward dynamics model to capture both the aleatoric and epistemic uncertainties of the model. 
Following prior work using neural networks as dynamics models~\cite{chua2018deep, nagabandi2020deep}, we capture the aleatoric uncertainty using output neurons which parameterize a normal distribution: $\dynamicsmodel(\cdot |\vec s_t, \vec a_t) = \mathcal{N}(\mu_{\vec \theta}( \vec s_t, \vec a_t), \Sigma_{\vec \theta}(\vec s_t, \vec a_t))$. 
We also capture the epistemic uncertainty through the use of ensembles~\cite{chua2018deep} by taking the average disagreement between the models in the ensemble.

The model is trained incrementally by maximizing the log-likelihood based on the transition data stored in the replay buffer.
The transition data is collected from experience of evaluations performed during the skill discovery stage with QD. 
Learning dynamics models in this self-supervised manner does not incur any additional cost to obtain data, and allows us to make use of the dense training signals from each state transition. 
Our approach differs from M-QD which considers only the skills present in the repertoire as a dataset to update its model. 
Conversely, \algoname{} considers the transitions from every evaluated skill, including those not added to the repertoire.

\subsection{Performing QD Exploration in Imagination} \label{sec:methods-qd_exploration_imag}

\algoname{} uses the imagined trajectories introduced in the previous section to build an imagined repertoire $\archivesynthetic$.
This imagined skill repertoire works in the same way as a standard skill repertoire but only contains imagined skills (i.e., that have been exclusively evaluated with the dynamics model).
From imagined trajectories, we can obtain imagined skill descriptors $\smash{\widetilde{\skilldescriptor}}$ and imagined returns~$\smash{\widetilde{\reward}}$.
Following the addition condition (section~\ref{sec:background}), if $\smash{\widetilde{\skilldescriptor}}$ is novel enough, or if $\smash{\widetilde{\reward}}$ is high enough, the policies are added to the imagined repertoire $\archivesynthetic$.

We keep performing QD exploration in imagination from the imagined skill repertoire $\archivesynthetic$ until a stopping criterion is met.
The nature of this stopping criterion may be adapted according to our needs.
In the simplest case, we define a fixed number of iterations to perform QD in imagination.
For example, if we want to use this framework as a simple surrogate model, the stopping criterion is set to one iteration of imagined rollouts. 
Alternatively, we can set this condition to be dependant on the state of the imagined repertoire (i.e. repertoire size, average return, number of skills added, etc.). 

\subsection{Acting in the Environment} \label{sec:methods-acting_in_env}

After performing QD exploration in imagination, skills are selected from the imagined repertoire $\archivesynthetic$ to be evaluated in the environment. 
There are several possible ways to perform this selection.
For instance, we can choose to select all policies that have been successfully added to the imagined repertoire, as in~\ref{sec:exps-sample_eff}. 
Alternatively, we can take into account the uncertainty in model predictions for the policies in the imagined repertoire: policies with low model uncertainty are likely to work in the environment, as in~\ref{sec:exps-continual}.
Transitions from evaluated policies are stored in the replay buffer.

Once these policies are evaluated, we update the repertoire $\archivereal$ following its addition condition (see section~\ref{sec:background}).
As soon as $\archivereal$ is updated, the two repertoires are synced: the content of the imagined repertoire $\archivesynthetic$ is replaced with the content from $\archivereal$.
Finally, the dynamics model $\dynamicsmodel$ is trained with the updated replay buffer after all selected skills have been evaluated.
This way, we ensure our imagined repertoire and the dynamics model are as close to the environment as possible. 

\section{Experiments}
Our experiments address three questions: 
(1) Are dynamics models more effective surrogate models than direct models for skill discovery using QD?
(2) Does the \algoname{} framework with an imagined repertoire allow for zero-shot and few-shot acquisition of new skills?
(3) Given a limited number of evaluations, does \algoname{} outperform vanilla QD on a long horizon real-world navigation task?
In our experiments (\ref{sec:exps-sample_eff} and \ref{sec:exps-continual}), we perform skill discovery in simulation. The resulting skill repertoires learnt are then used to perform navigation and damage recovery in the real-world (\ref{sec:navigation}).

\subsection{Environment} \label{sec:exps-envs}
\textbf{Hexapod Locomotion.} For our experiments, we use an  open-sourced 18-DoF hexapod robot (see Figure \ref{fig:images-hexapod-trajectory-damaged}). 
The goal for the robot is to learn primitive locomotion skills that enable it to move in every direction. We call this the omni-directional skill repertoire. To do this, we use a 2-D skill descriptor defined by the final $(x,y)$ position reached by the robot after an episode of three seconds. The reward for this skill repertoire encourages the robot to follow a circular trajectory~\cite{cully2013behavioral, chatzilygeroudis2018reset} defined by $R = |\alpha_i-\alpha_d|$, where $\alpha_i$ is the robot's final orientation and $\alpha_d$ the tangent of the desired circular trajectory. The resulting skill repertoire consists of skills as shown in Figure~\ref{fig:hexa_omni_ablation}A.

We define a primitive policy as a parametric low-level controller, as used in~\cite{cully2015robots, chatzilygeroudis2018reset}. Each joint angle follows a periodic function parameterized by the amplitude, phase and duty cycle of a sinusoid wave. The angular position of the third joint of each leg is opposite to the second leg, leading to 12 independently controlled joints. In total, a movement primitive consists of 36 parameters, $\vec \phi$. We use this policy representation for its simplicity in showing the performance of the overall algorithm. It also injects periodic movement priors which are useful for locomotion. However, more complex policies with a larger number of parameters such as deep neural networks can also be used in the future with our method~\cite{colas2020scaling, nilsson:hal-03135723}.

\subsection{Sample Efficiency} \label{sec:exps-sample_eff}

\begin{figure}[t]
\centering
	\includegraphics[width=0.49\textwidth]{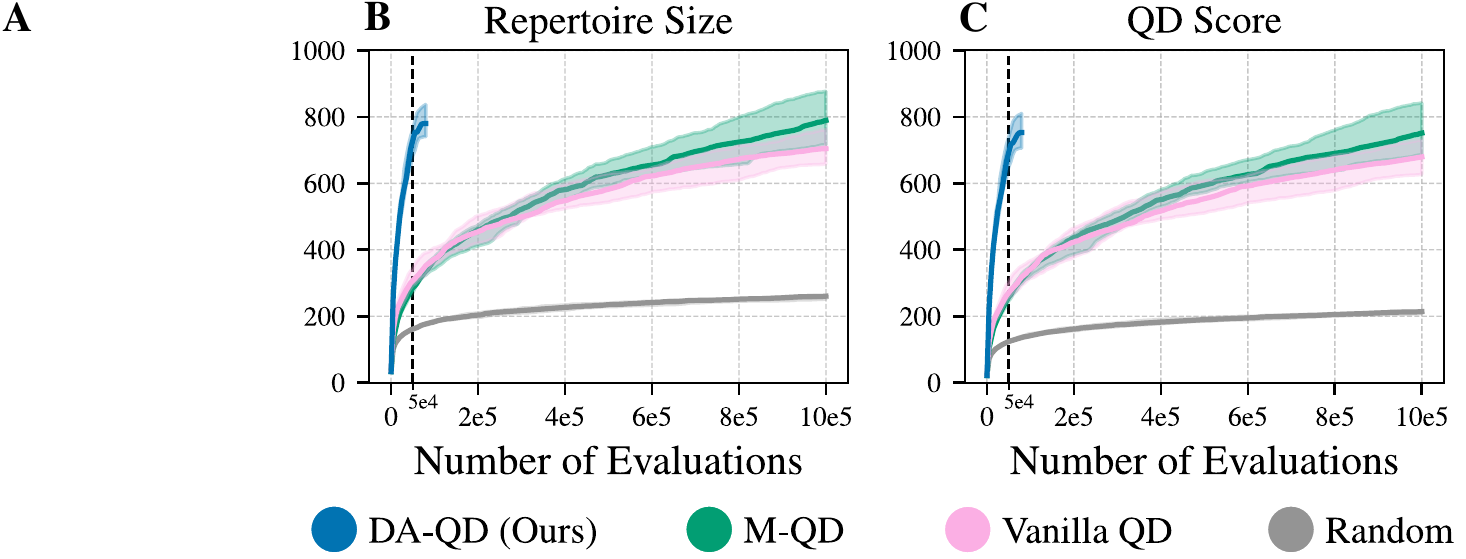}
\caption{%
(\textbf{A}) Illustration of the omni-directional skill repertoire for the hexapod;
(\textbf{B} and \textbf{C}) Evolution of repertoire size and QD score for the variants under study.
Each curve presents the evolution of the median performance from 10 replications.
The shaded areas represent the associated interquartile range.
For \algoname{}, we report the size and QD score of the repertoire $\archivereal$. 
%
%
}\label{fig:hexa_omni_ablation}
\vspace{-5mm}
\end{figure}

We first show that \algoname{} improves the sample efficiency of skill discovery in QD algorithms. 

\textbf{Baselines and Metrics.} 
We compare \algoname{} against random exploration, Vanilla QD and M-QD~\cite{keller2020model} (QD with direct model) as baselines.
We evaluate the performance of these algorithms using two metrics: (1) the \textit{repertoire size}, which measures the coverage and diversity of the skill repertoire; and (2) the \textit{QD-score}~\cite{pugh2016quality}, which measures the quality of the skills in the repertoire.
The QD-score is computed as the sum of the returns of all the policies in the repertoire. 

\textbf{Results.}
Figure \ref{fig:hexa_omni_ablation} shows curves for the progress of the metrics during skill discovery on the hexapod environment. With a median repertoire size and QD-score of $738$ and $694$ after $5\times10^4$ evaluations, \algoname{} outperforms vanilla QD with $10^6$ evaluations on both metrics. \algoname{} also performs on par and is equivalent to M-QD with a median repertoire size of $705$ and a QD score of $680$ after $10^6$ evaluations. However, \algoname{} achieves a significant increase ($\sim$20 times) in sample efficiency.
As \algoname{} approaches convergence, the number of skills added to the imagined repertoire significantly decreases. We stop the algorithm as soon as less than one policy is added to the imagined repertoire in one iteration. 

These results demonstrate that our dynamics model is a better surrogate model than the direct model (as used in M-QD) and leads to high data efficiency. We hypothesize that this is due to the dense training signals from each state transition. We can leverage a much larger dataset to train the dynamics model in comparison to learning a direct model, where one data point corresponds to one evaluation. Additionally, learning a mapping from parameter space to the skill space is potentially more difficult than modeling the transition dynamics. The mapping required to be learnt by the direct model also makes the performance sensitive to the choice of parameter space used.  

\subsection{Continual Acquisition of Skills} \label{sec:exps-continual}

One of the benefits of learning a dynamics model is that we can re-use the model for another task. 
Given a previously learnt dynamics model $\dynamicsmodel$, we can fully perform QD skill discovery in imagination to acquire additional skill repertoires.
As opposed to modeling the next state distribution like in our dynamics model, the direct model from M-QD directly learns a mapping from the parameter space to the skill descriptor and reward space. 
This makes the direct model specific to the skill repertoire it was trained on and cannot be re-purposed to learn another skill.


%
%

\textbf{Task and Metrics.} Using the same hexapod environment as above and the corresponding learnt dynamics model from \ref{sec:exps-sample_eff}, we now define a different skill descriptor that aims to find many diverse ways to walk forward as fast as possible (details in~\cite{cully2015robots}).
%
%
%
This skill descriptor characterises changes in the angular position of the robot when walking.
This leads to skills with diverse forward locomotion gaits. 
%
To assess the quality of the produced skill repertoires, we report the best return, mean return and size of the skill repertoires.
We show that \algoname{} is capable of (1) acquiring a new skill repertoire, and (2) learning skills in a zero-shot and few-shot manner for this new task.

\subsubsection{Acquisition of New Skill Repertoires}
\label{sec:exps-acquistion_new_rep}
%
After performing QD in imagination for 100,000 imagined rollouts (without any evaluations in the environment), we select all the policies present in the fully imagined repertoire $\archivesynthetic$. We evaluate all of them and attempt to add them to the skill repertoire $\archivereal$.
%
%
%

\textbf{Baselines.} 
As an equivalent baseline, we consider repertoires generated using vanilla QD with the same number of evaluations: $3{,}000$. This number corresponds to the approximate size of the imagined skill repertoires obtained by \algoname{} and which are evaluated to form its final repertoires.
As an upper bound, we also compare our results with skill repertoires generated using vanilla QD with $100{,}000$ evaluations, which corresponds to the number of imagined rollouts performed in \algoname{}. 


\textbf{Results.} 
Naturally, due to the inaccuracies of the model $\dynamicsmodel$, the final skill space coverage (repertoire size) of \algoname{} is lower than its imagined repertoire size (see Table~\ref{table:continual-results}).
Yet, skills discovered by \algoname{} in imagination significantly outperform the equivalent vanilla QD variant with the same number of evaluations both in terms of performance and diversity. 
While the skill repertoire size is smaller compared to the upper baseline which uses $10^5$ (${\sim}30\times$ more) evaluations, \algoname{} achieves similar best return and better average return.

\subsubsection{Zero-shot and Few-shot Acquisition of Skills}
\label{sec:exps-zero_few_shot}
Instead of evaluating all the skills of the imagined repertoire, like in the previous section, we can evaluate only a few to obtain new skills in a zero-shot or few-shot manner. The model uncertainty (\ref{sec:methods-dyn_model}) is used to select from the imagined repertoire skills that are most likely to succeed when executed. We select 20 skills with the lowest model disagreement. From this set, the skill with the best expected return is selected to be executed. We use this selection procedure to evaluate zero-shot performance. 
When more trials are permitted, the $n$ best skills in the set can be executed and evaluated for $n$-shot performance. 
We only report zero-shot and 20-shot results as preliminary experiments show that the best return increases as the number of taken trials increases.

\textbf{Results.} Table~\ref{table:continual-results} shows that using the above selection procedure enables \algoname{} to obtain skills that successfully solve the new task with minimal or no prior environment interaction. 
\algoname{} (20-shot) performs very similarly to the best policy found by Vanilla QD (Equivalent) while requiring 150 times fewer evaluations. 
%
These few-shot learning capabilities were not possible using prior QD methods.

\subsection{Navigation and Damage Recovery in Real-World} 
\label{sec:navigation}

\begin{figure}[t]
\centering
	\includegraphics[width=0.49\textwidth]{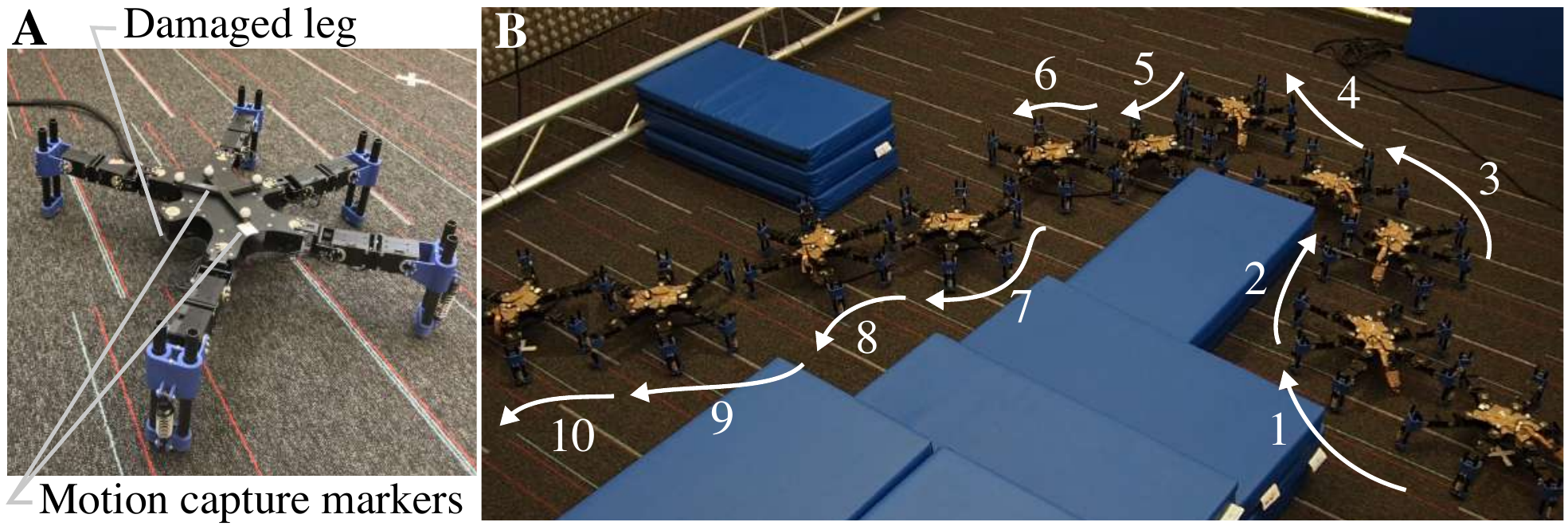}
\centering
\caption{%
(\textbf{A}) Picture of the damaged robot;
(\textbf{B}) Example trajectory obtained in the real world for \algoname{}, where 10 skills are executed to reach the goal. 
}
\vspace{-5mm}
\label{fig:images-hexapod-trajectory-damaged}
\end{figure} 

Lastly, we show how \algoname{} can be useful in a real-world long-horizon navigation task, and that the increase in sample-efficiency during skill discovery does not come at the cost of planning and damage recovery capabilities.

\begin{figure}[t]
\centering
	\includegraphics[width=0.49\textwidth]{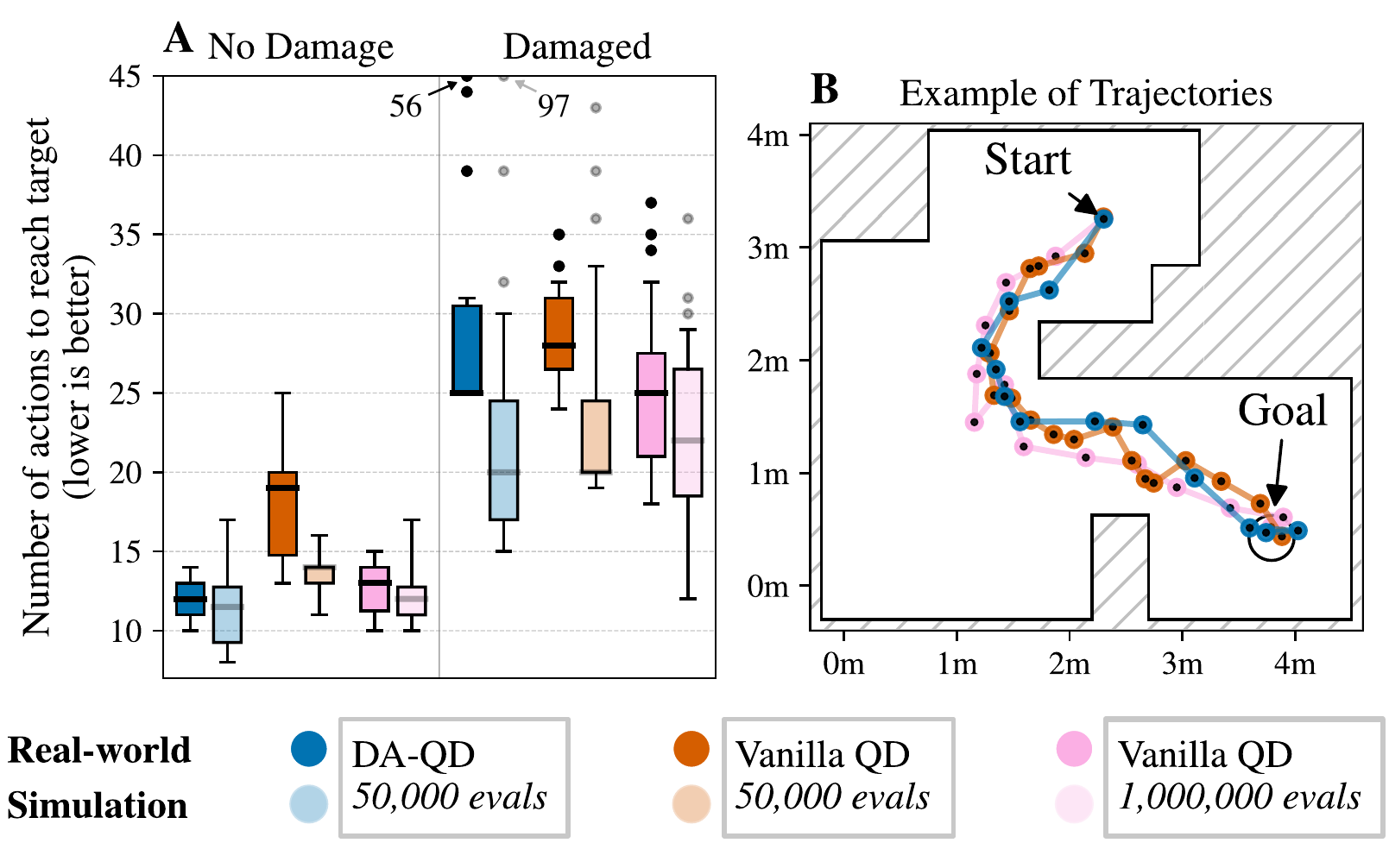}
\centering
\vspace{-8mm}
\caption{%
(\textbf{A}) Number of actions executed before reaching the goal in the \textit{No Damage} and \textit{Damaged} scenarios. 
Results are provided for both the real-world and the simulation settings. 
Each box-plot illustrates the distribution obtained from 10 replications. 
%
%
For each variant in the damaged scenario,  the three worst-performing repertoires are represented as outlier points.
(\textbf{B}) Example of real-world trajectories (obtaining median scores in \textbf{A}).
%
}
\vspace{-5mm}
\label{fig:rte-results}
\end{figure}

\textbf{Task.} We set up a navigation task in which the robot has to navigate through a maze to a goal position using the skill repertoire $\archivereal$ discovered from simulation (via \algoname{} or any other QD algorithm).
We consider two different scenarios: (1) a \textit{no-damage} scenario where the robot is intact, and (2) a \textit{damaged} scenario in which one of the legs of the robot is removed (Fig.~\ref{fig:images-hexapod-trajectory-damaged}).
The damaged scenario requires the robot to adapt to its damaged condition and solve the task at the same time. 
%
It is important to note that skill discovery in simulation is performed with a fully intact robot, and the robot has no knowledge of its damaged condition when deployed in the real world. 
%
%
For our experiments, we provide the robot with a 2-D layout of the map it has to navigate as seen in Figure \ref{fig:rte-results}~B. 
We use the Reset-free Trial and Error (RTE)~\cite{chatzilygeroudis2018reset} algorithm to solve this navigation task.

\textbf{Skill-space Planning and Adaptation via RTE.} The RTE algorithm relies on a skill repertoire $\archivereal$ to (1) perform planning to solve long-horizon navigation tasks in the real world, and (2) adapt to sim-to-real differences and potential damaged conditions.
To do so, RTE utilises Gaussian Processes (GPs) to predict the difference between the outcome of a skill in simulation and in the real world.
%
The GP predictions are then used by a Monte-Carlo Tree Search (MCTS) planner to choose the best skill to execute from $\archivereal$.
After each executed skill, the resulting position and orientation of the robot are measured (via a motion-capture system), and the GP predictions are updated accordingly. 
More details about RTE are provided in the work of Chatzilygeroudis et al. \cite{chatzilygeroudis2018reset}. 

\textbf{Metric.} For each experimental run, we evaluate the total number of repertoire skills that have been executed by the robot to reach the goal.
A low number of executed skills reflects the high quality and diversity of the repertoire.

\begin{table}[t]
\footnotesize
\centering
\caption{Continual Skill Acquisition Results}
\vspace{-3mm}
\begin{threeparttable}
\begin{tabular}{ lrrrr } 
 \toprule
 Algorithm & Evaluations & \begin{tabular}{@{}r@{}}
                   Best\\
                   Return\\
                 \end{tabular}  & \begin{tabular}{@{}r@{}}
                   Mean\\
                   Return\\
                 \end{tabular}  & \begin{tabular}{@{}r@{}}
                   Repertoire\\
                   Size\\
                 \end{tabular}  \\
 \midrule
 \algoname{} & $2{,}705\tnote{1}$ & $0.471$ & $0.191$ & $838$\\ 
QD (Equivalent) & $3{,}000$ & $0.353$ & $0.068 $ & $662 $ \\
QD (Upp. Baseline) & $100{,}000$ & $0.497$ & $0.177$ & $3{,}225$ \\ 
 \midrule
 \algoname{} (20-shot) & 20 & $0.33$ & $0.11$ & - \\ 
 \algoname{} (0-shot) & 1 & $0.16$ & - & - \\ 
 \bottomrule
\end{tabular}
\begin{tablenotes}
     \item For each variant, we report the median result from 10 replications.
     \item[1] The number of evaluations corresponds to the imagined repertoire size.
 \end{tablenotes}
\end{threeparttable}
\vspace{-5mm}
\label{table:continual-results}
\end{table}

\textbf{Baselines.} To show the effectiveness of using \algoname{}, we compare skill repertoires generated with \algoname{} with just 50,000 evaluations to some baselines for these tasks. 
As an equivalent baseline, we use vanilla QD to generate skill repertoires with only $50{,}000$ evaluations. 
As an upper bound baseline, we do the same but with vanilla QD using $10^6$ evaluations.
For each variant, we generate 10 different repertoires from independent replications and use RTE with each of those repertoires.
This results in 10 replications per variant, both in the real world and in simulation.

\textbf{Results.} Although it uses 20 times fewer evaluations, the results obtained by \algoname{} are mostly equivalent to the ones from the upper baseline having access to $10^6$ evaluations (Fig.~\ref{fig:rte-results}).
This similarity is noticeable in the no-damage scenario (Fig.~\ref{fig:rte-results}~A), and also in simulation for the damaged scenario.
Thus, \algoname{} learns a repertoire that is as efficient for long-horizon task solving as the repertoires produced with $20$ times more evaluations.

In the no-damage scenario, \algoname{} outperforms the equivalent vanilla QD baseline, both in the real world and in simulation (Fig.~\ref{fig:rte-results}~A).
As shown in Figure \ref{fig:rte-results}~B, \algoname{} requires fewer skills to reach the goal because its displacements tend to be larger. This corresponds to results from \ref{sec:exps-sample_eff} which show that \algoname{} discovers skill repertoires that have greater coverage and performance over baselines.
In the damaged scenario, some repertoires learnt in simulation heavily rely on the damaged leg to solve the task. 
%
As RTE does not provide any mechanism to handle this issue, those repertoires performed poorly.
The resulting outliers make the comparison challenging between the different variants.
If we compare the performance of those distributions without those outlying performances, \algoname{} performs slightly better than its equivalent QD algorithm (Fig.~\ref{fig:rte-results}~A).
Given a budget of $5\times 10^4$ evaluations, \algoname{} learns a repertoire that is more beneficial for long-horizon task solving than those obtained from Vanilla QD algorithms.

\section{Discussion and Conclusion} \label{sec:discussion}

\textbf{Summary.} We introduced a novel algorithm, \algoname{}, for sample efficient discovery of diverse skills by combining a dynamics model and imagined skill repertoire with Quality-Diversity. 
Thanks to this, \algoname{} is about 20 times more data-efficient than previous methods. 
\algoname{} also demonstrated zero-shot and few-shot skill acquisition which was not possible before this using prior QD methods.
We further show robots using \algoname{} are capable of long-horizon navigation and damage adaptation.

\textbf{Limitations.}
As we use deep neural networks as dynamics models, we also inherit their common limitations. Of concern to us, the model accuracy of longer horizon rollouts can impact the performance of \algoname{}. Future research can use latent dynamics models which work better for long-term predictions~\cite{hafner2019dream} and also allow DA-QD to scale to higher-dimensional and more complex domains. 
While effective for planning, the skill spaces explored are still low dimensional; this limits scaling to more complex skills. Future work will aim to combine DA-QD with existing approaches to skill descriptor learning~\cite{cully2019autonomous, pere2018unsupervised, paolo2020unsupervised} using unsupervised representation learning methods like autoencoders.

%
%
%
%

\addtolength{\textheight}{-2cm}   




\section*{ACKNOWLEDGMENT}

This work was supported by the Engineering and Physical Sciences Research Council (EPSRC) grant EP/V006673/1 project REcoVER.


\bibliographystyle{IEEEtran} 
\bibliography{biblio}

\end{document}